\pdfoutput=1

\documentclass[11pt]{article}

\usepackage{ACL2023}

\usepackage{times}
\usepackage{latexsym}

\usepackage[T1]{fontenc}

\usepackage[utf8]{inputenc}

\usepackage{microtype}

\usepackage{inconsolata}
\usepackage{graphicx}
\usepackage{subcaption}
\usepackage{multirow}
\usepackage{tabularx}

\title{Enhancing Translation for Indigenous Languages: Experiments with Multilingual Models}

\author{
\normalsize Atnafu Lambebo Tonja$^{1}$, Hellina Hailu Nigatu$^{2}$, Olga Kolesnikova$^{1}$, \\
\textbf{\normalsize  Grigori Sidorov$^{1}$, Alexander Gelbukh$^{1}$, Jugal Kalita$^{3}$}  \\
\footnotesize
$^1$Instituto Politécnico Nacional (IPN), Mexico,\\
\footnotesize
$^2$University of California, Berkeley, USA, \\
\footnotesize
$^3$University of Colorado, Colorado Springs, USA \\
}

\begin{document}
\maketitle
\begin{abstract}
This paper describes CIC NLP's submission to the AmericasNLP 2023 Shared Task on machine translation systems for indigenous languages of the Americas. We present the system descriptions for three methods. We used two multilingual models, namely M2M-100 and mBART50, and one bilingual (one-to-one) --- Helsinki NLP Spanish-English translation model, and experimented with different transfer learning setups. We experimented with 11 languages from America and report the setups we used as well as the results we achieved. Overall, the mBART setup was able to improve upon the baseline for three out of the eleven languages. 
\end{abstract}

\section{Introduction}
While machine translation systems have shown commendable performance in recent years, the performance is lagging for low-resource languages \cite{pmlr-v176-hadgu22a, app13021201}. Since low-resource languages suffer from a lack of sufficient data \cite{siddhant20221000, 10.1162/coli_a_00446}, most models and methods that are developed for high-resource languages do not work well in low-resource settings. Additionally, low-resource languages are linguistically diverse and have divergent properties from the mainstream languages in NLP studies \cite{shared_task_2021}. 

Though low-resource languages lack sufficient data to train large models, some such languages still have a large number of native speakers \cite{shared_task_2021}. While the availability of language technologies such as machine translation systems can be helpful for such linguistic communities, they could also bring harm and exposure to exploitation \cite{hovy-spruit-2016-social}. Borrowing from human-computer interaction (HCI) studies \cite{10.1145/3173574.3173818}, we want to acknowledge our belief that low-resource language speakers should be empowered to create technologies that benefit their communities. Many indigenous communities have community-rooted efforts for preserving their languages and building language technologies for their communities \footnote{\url{https://papareo.nz/}} and we hope that methods from Shared Tasks like this will contribute to their efforts.

Improving machine translation systems for low-resource languages is an active research area and different approaches \cite{zoph2016transfer,karakanta2018neural,ortega2020neural,goyal2020efficient,tonja2022improving,imankulova2017improving} have been to improve the performance of systems geared forward low-resource languages.  We participated in the AmericasNLP 2023 Shared Task in hopes of contributing new approaches for low-resource machine translation that are likely to be helpful for community members interested in developing and adapting these technologies for their languages. 

In recent years, large pre-trained models have been used for downstream NLP tasks, including machine translation \cite{brants-etal-2007-large} because of the higher performance in downstream tasks compared to traditional approaches \cite{han2021pre}. One trend is to use these pre-trained models and fine-tune them on smaller data sets for specific tasks \cite{sun2019fine}. This method has shown promising results in downstream NLP tasks for languages with low or limited resources \cite{tars2022teaching,zhao2022improving}. In our experiments, we used multilingual and bilingual models and employed different fine-tuning strategies for the eleven languages in the 2023 Shared Task \cite{ebrahimi-etal-2023-findings}. 

In this paper, we describe the system setups we used and the results we obtained from our experiments. One of our systems improves upon the baseline for three languages. We also reflect on the setups we experimented with but ended up not submitting in hopes that future work could improve upon them. 


\section{Languages and Datasets}
In this section, we present the languages and datasets used in our shared task submission. Table \ref{tab:datasize} provides an overview of the languages, their linguistic families, and the numbers of parallel sentences.
\begin{table}[h!]
\tiny
\begin{tabular}{llllll}
\hline
\textbf{Language} & \textbf{ISO} & \textbf{Family} & \textbf{Train} &
\textbf{Dev} & \textbf{Test} \\ \hline
Aymara         & aym & Aymaran      & 6,531   & 996 & 1,003 \\
Bribri         & bzd & Chibchan         & 7,508   & 996 & 1,003 \\
Asháninka      & cni & Arawak          & 3,883   & 883 & 1002 \\
Chatino        & czn & Zapotecan        & 357    & 499 & 1,000 \\
Guarani        & gn  & Tupi-Guarani  & 26,032  & 995 & 1,003 \\
Wixarika       & hch & Uto-Aztecan      & 8,966   & 994 & 1003 \\
Nahuatl        & nah & Uto-Aztecan  & 410,000    & 672 & 996  \\
Hñähñu         & oto & Oto-Manguean    & 4,889   & 599 & 1,001 \\
Quechua        & quy & Quechuan     & 125,008 & 996 & 1,003 \\
Shipibo-Konibo & shp & Panoan        & 14,592  & 996 & 1,002 \\ 
Rarámuri & tar & Uto-Aztecan  & 14721 & 995 & 1002 \\ \hline
\end{tabular}
\caption{This table provides information about the languages with which we experimented including ISO language code and language family as well as the number of sentences in training, development, and test sets for each language.}
\label{tab:datasize}
\end{table}
\paragraph{Aymara} is an Aymaran language spoken by the Aymara people of the Bolivian Andes. It is one of only a handful of Native American languages with over one million speakers \cite{homola2012building}. Aymara, along with Spanish and Quechua, is an official language in Bolivia and Peru. The data for the Aymara-Spanish come from the Global Voices \cite{tiedemann-2012-parallel}.

\paragraph{Bribri} The Bribri language is spoken in Southern Costa Rica. Bribri has two major orthographies: Jara\footnote{https://www.lenguabribri.com/se-tt\%C3\%B6-bribri-ie-hablemos-en-bribri} and Constenla\footnote{https://editorial.ucr.ac.cr/index.php} and the writing is not standardized which results in spelling variations across documents. In this case, the sentences use an intermediate representation to unify existing orthographies. The Bribri-Spanish data \cite{back_translation_2020} came from six different sources. 

\paragraph{Asháninka} Asháninka is an Arawakan language spoken by the Asháninka people of Peru and Acre, Brazil\footnote{https://www.everyculture.com/wc/Norway-to-Russia/Ash-ninka.html}. It is primarily spoken in the Satipo Province located in the Amazon forest. The parallel data for Asháninka-Spanish come mainly from three sources \cite{cushimariano:prel:08, ortega-etal-2020-overcoming, mihas:anaani:11} and translations by Richard Castro. 

\paragraph{Chatino} Chatino is a group of indigenous Mesoamerican languages. These languages are a branch of the Zapotecan family within the Oto-Manguean language family. They are natively spoken by 45,000 Chatino people \cite{cruz2006sandhi} whose communities are located in the southern portion of the Mexican state of Oaxaca. The parallel data for Chatino-Spanish can be accessed here\footnote{\url{https://scholarworks.iu.edu/dspace/handle/2022/21028}}.

\paragraph{Guarani} Guarani is a South American language that belongs to the Tupi-Guarani family \cite{britton2005guarani} of the Tupian languages. It is one of the official languages of Paraguay (along with Spanish), where it is spoken by the majority of the population, and where half of the rural population are monolingual speakers of the language \cite{mortimer2006guarani}.

\paragraph{Wixarika} Wixarika is an indigenous language of Mexico that belongs to the Uto-Aztecan language family \cite{de2003ley}. It is spoken by the ethnic group widely known as the Huichol (self-designation Wixaritari), whose mountainous territory extends over portions of the Mexican states of Jalisco, San Luis Potosí, Nayarit, Zacatecas, and Durango, but mostly in Jalisco. United States: La Habra, California; Houston, Texas. 

\paragraph{Nahuatl} Nahuatl is a Uto-Aztecan language and was spoken by the Aztec and Toltec civilizations of Mexico\footnote{www.elalliance.org/languages/nahuatl}. The Nahuatl language has no standard orthography and has wide dialectical variations \cite{shared_task_2021}.  

\paragraph{Hñähñu} Hñähñu, also known as Otomí, belongs to the Oto-Pamean family and lived in central Mexico for many centuries \cite{Lastra_2001}. Otomí is a tonal language with a Subject-Verb-Object (SVO) word order \cite{ebrahimi2022americasnli}. It is spoken in several states across Mexico.

\paragraph{Quechua} The Quechua-Spanish data \cite{agic-vulic-2019-jw300, tiedemann-2012-parallel} has three different sources: the Jehova's Witnesses texts, the Peru Minister of Education, and dictionary entries and samples collected by Diego Huarcaya. The Quechua language, also known as Runasimi is spoken in Peru and is the most widely spoken pre-Columbian language family of the Americas \cite{ebrahimi2022americasnli}. 

\begin{figure*}[h!]
  \centering
  \begin{subfigure}[b]{0.3\textwidth}
    \centering
    \includegraphics[width=\textwidth]{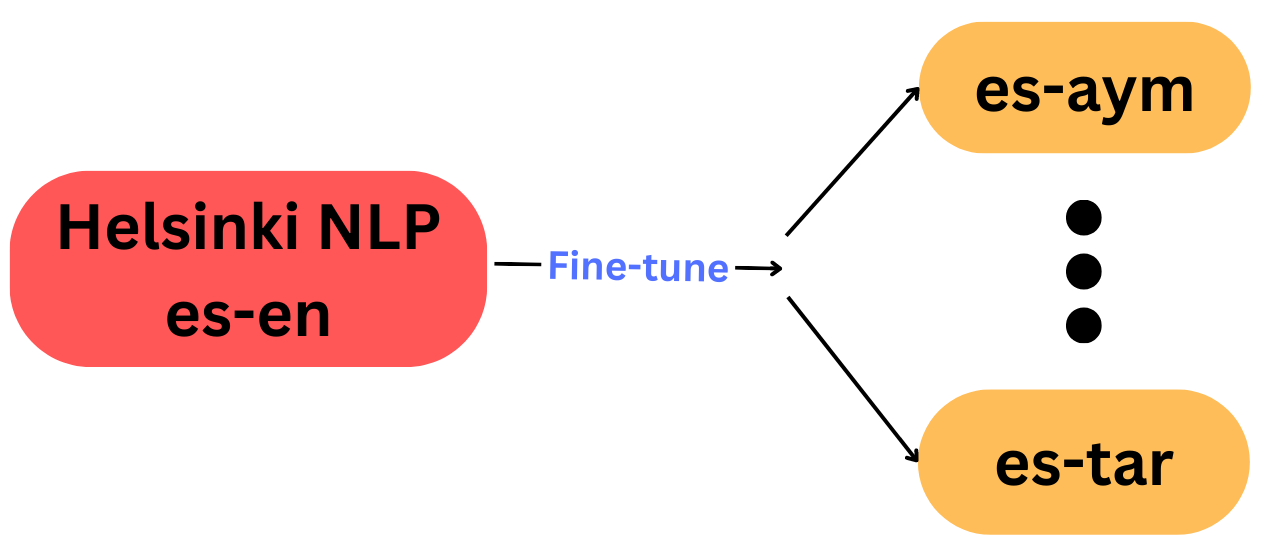}
    \caption{Spanish-English (Helsinki) model}
    \label{fig:figure1}
  \end{subfigure}
  \hfill
  \begin{subfigure}[b]{0.3\textwidth}
    \centering
    \includegraphics[width=\textwidth]{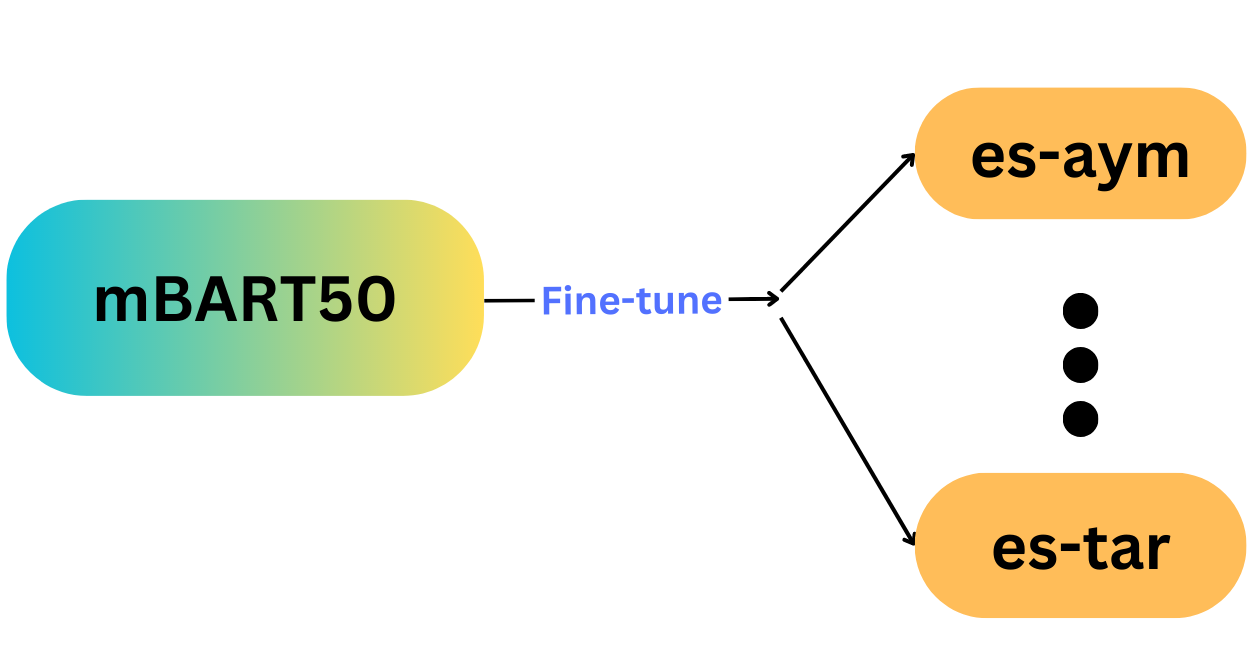}
    \caption{mBART50 model }
    \label{fig:figure2}
  \end{subfigure}
  \hfill
  \begin{subfigure}[b]{0.3\textwidth}
    \centering
    \includegraphics[width=\textwidth]{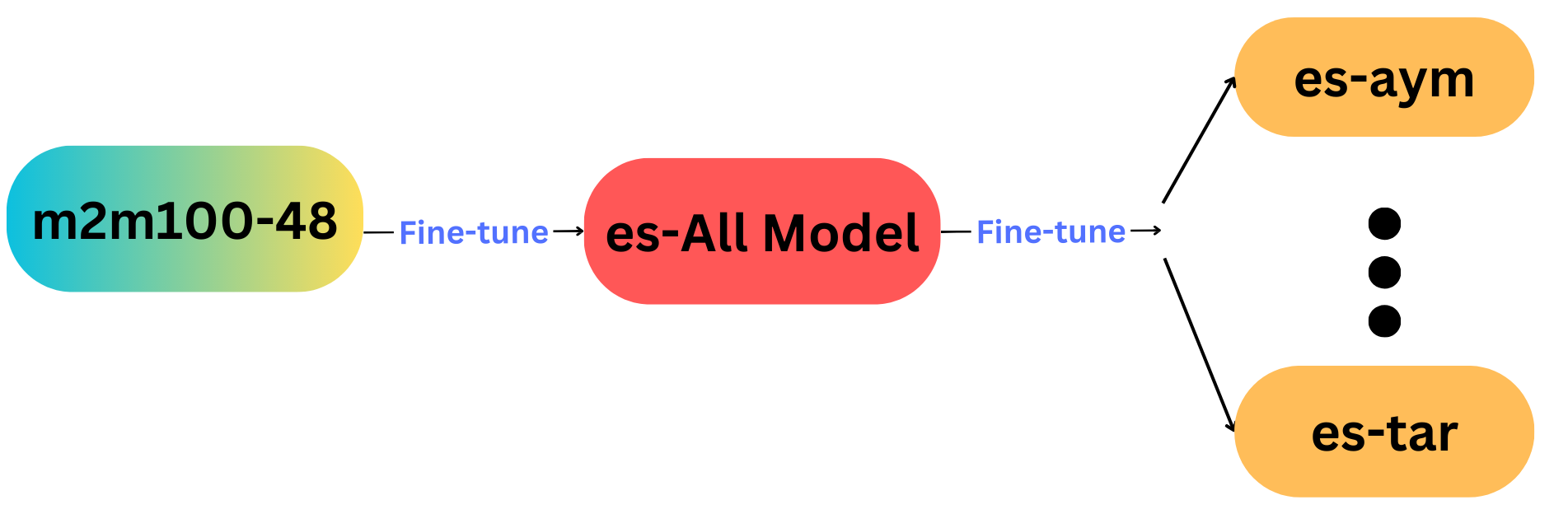}
    \caption{m2m100-48$_{inter}$  model}
    \label{fig:figure3}
  \end{subfigure}
  \caption{Experiments on \textbf{(a)} fine-tuning bilingual model, \textbf{(b)} and \textbf{(c)} fine-tuning multilingual models. For \textbf{(a)} we fine-tuned the bilingual Spanish-English model on Spanish-Indigenous pairs, for \textbf{(b)} we fine-tuned the multilingual mBART50 model on Spanish-Indigenous pairs, and for \textbf{(c)} we fine-tuned the multilingual m2m100-48 model first on Spanish-All to produce m2m100-48$_{inter}$  model and then fine-tuned the m2m100-48$_{inter}$ model on Spanish-Indigenous pairs.}
  \label{fig:all-model}
\end{figure*}
\paragraph{Shipibo-Konibo} Shipibo-Konibo - Spanish data \cite{montoya-etal-2019-continuous, galarreta-etal-2017-corpus} come from three different sources: samples from flashcards translated to Shipibo-Konibo, sentences translated from books for bilingual education, and dictionary entries. 

\paragraph{Rarámuri}  Rarámuri, also known as Tarahumara is a Uto-Azetcan language spoken in Northern Mexico \cite{gabe_2017}. Rarámuri is a polysynthetic and agglutinative language spoken mainly in the Sierra Madre Occidental region of Mexico \cite{ebrahimi2022americasnli}.

\begin{table*}[t]
\small
\centering
\begin{tabular}{l|l|rllllllllll|l}
\hline
\textbf{Data} &
  \textbf{Model} &
  \multicolumn{1}{l}{\textbf{aym}} &
  \textbf{bzd} & \textbf{cni} &  \textbf{czn} &  \textbf{gn} & \textbf{hch} & \textbf{nah} &   \textbf{oto} &  \textbf{quy} & \textbf{shp} &\textbf{tar} &   Average \\ \hline
\multicolumn{2}{l}{\textbf{Baseline}} &
  \multicolumn{1}{l}{28.3} & 16.5 &   25.8 &   - &   33.6 & 30.4 &  26.6 &  14.7 &  34.3 &  32.9 &  18.4 &  - \\ \hline \hline
 &  M1 &  12.25 &  20.3 &   26.65 &  - &  23.83 &  11.09 & 29.55 &  6.57 &  35.04 & 20.99 & 14.12 &  20.03 \\
 & M2 & 
 20.65 & 18.59 &  20.63 &  - & 20.40 & 12.7 &  18.66 &  10.17 & 33.53 &21.03 & 13.54 &18.99 \\
 & M3 &
   14.70 & 19.9 & 25.62 &- & 23.62 & 11.82 &  29.94 & 7.94 &35.3 &21.32 &14.19 &\textbf{20.43} \\
\multirow{-4}{*}{\textbf{Dev}} &
  M4 & 20.89 &  12.17  &  23.59 & - &   20.84 &  13.51 & 22.63 &  7.16 &  30.86 &  18.02 & 12.60 &  18.22 \\ \hline
 & M2 &
  \textbf{19.05} & 19.90 & 23.50 &  14.41 &  19.35 &  12.05 &  21.88 &  \textbf{9.22} &  34.15 &  20.43 &  13.86 &  18.89 \\
 &  M3 &
 18.52 & \textbf{\underline{\underline{21.17}}} &  \textbf{\underline{\underline{25.85}}} &  \textbf{15.61} &  \textbf{21.75} &  13.88 &  \textbf{26.57} &   7.40 &
  \textbf{\underline{\underline{35.62}}} & \textbf{21.26} &  \textbf{14.87} &  \textbf{20.22} \\ 
\multirow{-3}{*}{\textbf{Test}} 
&  M4 &
 18.59 & 13.24 & 23.79 &  13.64 &  20.94 &  \textbf{14.67} &  22.60 &  7.28 &  32.75 &  18.13 &  12.07 &  17.97 \\ \hline
\end{tabular}
\caption{chrF2 scores for the three submissions, computed on the development and test sets. M1, M2, M3, and M4 represent M2M100-48, M2M100-48$_{inter}$, mBART50 and  Helsinki-NLP models respectively. The development set evaluations are used to select the best-performing model before working on submission data. The development set was not trained when evaluating the dev set, but we included the dev set during training for the final submission. The \textbf{\underline{\underline{bold}}} results show the models that out-preforms the baseline \cite{vazquez2021helsinki} results. The \textbf{bold} results show out-preforming models from our three model setups(excluding the baseline) for each individual language.}
\label{tab:results}
\end{table*}

\section{Models}
We experimented with two multilingual and one bilingual translation model with different transfer learning setups. We used M2M-100 and mBART50 for the multilingual experiment and the Helsinki-NLP Spanish-English model for the bilingual experiment. Figure \ref{fig:all-model} shows the models used in this experiment.

\subsection{Bilingual models}
For the bilingual model, as shown in Figure \ref{fig:figure1}, we use a publicly available Spanish - English\footnote{https://huggingface.co/Helsinki-NLP/opus-mt-es-en} pre-trained model from Huggingface\footnote{https://huggingface.co/} trained by Helsinki-NLP. The pre-trained MT models released by Helsinki-NLP are trained on OPUS, an open-source parallel corpus for covering 500 languages \cite{tiedemann-thottingal-2020-opus,tiedemann-2020-tatoeba}. This model is trained using the framework of Marian NMT \cite{mariannmt}. Each model has six self-attention layers in the encoder and decoder parts, and each layer has eight attention heads.

We used this model with the intention that the model trained with high-resource languages will improve the translation performance of low-resource indigenous languages when using a model trained with high-resource languages. We fine-tuned the Spanish-English model for each of the
Spanish-to-Indigenous language pairs.
\subsection{Multilingual models}
For multilingual models, we used the Many-to-Many multilingual translation model that can translate directly between any pair of 100 languages (M2M100) \cite{fan2021beyond} with 48M parameters and a sequence-to-sequence denoising auto-encoder pre-trained on large-scale monolingual corpora in 50 languages (mBART50) \cite{tang2020multilingual}.
We fine-tuned multilingual models in two ways:
\begin{enumerate}
    \item We fine-tuned two multilingual models on each Spanish-Indigenous language pair for 5 epochs and evaluated their performance using the development data before training the final submission system. As shown in Figure \ref{fig:figure2}, for the final system, we only fine-tuned mBART50 on Spanish-indigenous data based on the development set evaluation performance.
   \item Fine-tuning multilingual models first on the Spanish - All (mixture of all indigenous language data) dataset to produce an intermediate model and then fine-tuning the intermediate model for each of the Spanish-Indigenous language pairs as shown in Figure \ref{fig:figure3}. For this experiment, we combined all language pairs' training data to form a Spanish - all parallel corpus, and then we first fine-tuned m2m100-48 using a combined dataset for five epochs and saved the model, here referred to as m2m100-48$_{inter}$ model. We fine-tuned the m2m100-48$_{inter}$  model again on each Spanish-Indigenous language pair for another 5 epochs and evaluated the performance on the development set before training the final submission system.
\end{enumerate} 
\textbf{Evaluation} We used chrF2 \cite{popovic2017chrf} evaluation metric to evaluate our MT systems.
\section{Results}
We submitted three (two multilingual and one bilingual) systems, as shown in Table \ref{tab:results}, namely m2m100-48$_{inter}$, mBART50, and Helsinki-NLP. We included the dev set performance for all the models we trained before the final model to compare the results with the final model evaluated by using test set data. From the dev set result, it can be seen that fine-tuning the multilingual model on the Spanish-Indigenous language pair outperforms the fine-tuned result of the bilingual and m2m100-48$_{inter}$ models. From all the models evaluated using the dev set, mBART50 outperformed the others on average.

Our test results show comparable results when compared to the strongest baseline shared by the AmericasNLP 2023, and our model outperformed the baseline for Spanish-Bribri (es-bzd), Spanish-Asháninka (es-cni), and Spanish-Quechua (es-quy) pairs. Similarly, mBART50 outperformed the other models on average on the test set. 

\section{Conclusion}
In this work, we present the system descriptions and results for our submission to the 2023 AmericasNLP Shared Task on Machine Translation into Indigenous Languages. We used pre-trained models and tested different fine-tuning strategies for the eleven languages provided for the shared task. We used one bilingual (Helsinki NLP English-Spanish model) and two multilingual (M2M-100 and mBART50) models for our experiments. In addition to fine-tuning the individual languages' data, we concatenated the data from all eleven languages to create a Spanish-All dataset and fine-tuned the M2M-100 model before fine-tuning for the individual languages. Our mBAERT50 model beat the strong baseline in three languages. 
\section*{Acknowledgements}

\bibliography{anthology,custom}
\bibliographystyle{acl_natbib}




\end{document}